%% file: main.tex
\documentclass[10pt,twocolumn,letterpaper]{article}

\usepackage{wacv}              


\definecolor{wacvblue}{rgb}{0.21,0.49,0.74}
\usepackage[pagebackref,breaklinks,colorlinks,allcolors=wacvblue]{hyperref}
\usepackage{graphicx}
\usepackage{xspace}
\usepackage{xcolor}
\usepackage{array}
\usepackage{booktabs}     
\usepackage{multirow}     
\usepackage{tabularx}     
\usepackage{caption}      
\usepackage{ragged2e} 
\usepackage{arydshln} 
\usepackage{lipsum}   
\usepackage[accsupp]{axessibility}  

\usepackage[protrusion=true,expansion=true]{microtype}
\setlist[enumerate]{leftmargin=*, itemsep=1pt, parsep=0pt, topsep=1pt, partopsep=1pt}
\setlist[itemize]{leftmargin=*, itemsep=1pt, parsep=0pt,topsep=1pt, partopsep=1pt}

\def\wacvPaperID{***}
\def\confName{WACV}
\def\confYear{2026}

\title{MarineEval: Assessing the Marine Intelligence of Vision-Language Models}

\author{
Yuk-Kwan Wong \quad
Tuan-An To \quad
Jipeng Zhang \quad
Ziqiang Zheng\thanks{Corresponding author: zhengziqiang1@gmail.com} \quad
Sai-Kit Yeung\\
Hong Kong University of Science and Technology \\
\small
Project website: \url{https://marineeval.hkustvgd.com}
}

\newcommand{\datasetname}{MarineEval\xspace}
\newcommand{\modelCounts}{17\xspace}

\newcommand{\CTA}{\textit{C\&TA}\xspace}

\newcommand{\SR}{\textit{SR}\xspace}
\newcommand{\SC}{\textit{SC}\xspace}

\begin{document}

\let\oldtwocolumn\twocolumn
\renewcommand\twocolumn[1][]{%
    \oldtwocolumn[{#1}{
    \begin{center}
           \vspace{-1.5em}
           \includegraphics[width=\textwidth]{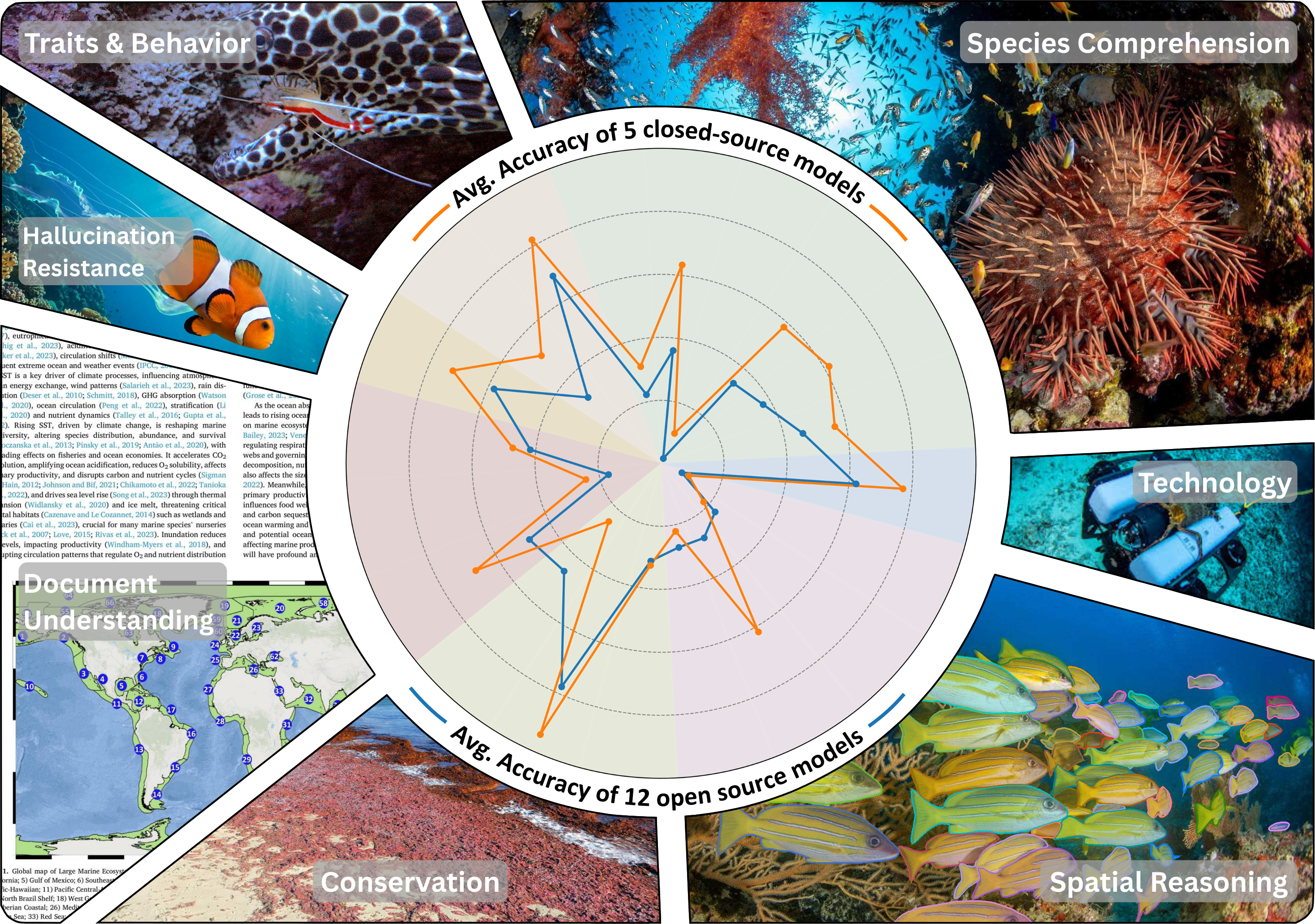}
           \captionof{figure}{We present MarineEval, the first large-scale dataset and benchmark to comprehensively assess the ability of existing vision language models (VLMs) on marine intelligence.
            }
           \label{fig:teaser}
        \end{center}
    }]
}

\maketitle
\input{sections/0_abstract}
\input{sections/1_introduction}

\input{sections/2_literature_review}
\input{sections/3_methodology}
\input{sections/4_experiments}
\input{sections/5_conclusion}

\input{sections/6_acknowledgement}

{
    \small
    \bibliographystyle{ieeenat_fullname}
    \bibliography{main}
}
\end{document}

%% file: sections/0_abstract.tex
\begin{abstract}
We have witnessed promising progress led by large language models (LLMs) and further vision language models (VLMs) in handling various queries as a general-purpose assistant. VLMs, as a bridge to connect the visual world and language corpus, receive both visual content and various text-only user instructions to generate corresponding responses. Though great success has been achieved by VLMs in various fields, in this work, we ask whether the existing VLMs can act as domain experts, accurately answering marine questions, which require significant domain expertise and address special domain challenges/requirements. To comprehensively evaluate the effectiveness and explore the boundary of existing VLMs, we construct the first large-scale marine VLM dataset and benchmark called MarineEval, with 2,000 image-based question-answering pairs. During our dataset construction, we ensure the diversity and coverage of the constructed data: 7 task dimensions and 20 capacity dimensions. The domain requirements are specially integrated into the data construction and further verified by the corresponding marine domain experts. We comprehensively benchmark \modelCounts existing VLMs on our MarineEval and also investigate the limitations of existing models in answering marine research questions. The experimental results reveal that existing VLMs cannot effectively answer the domain-specific questions, and there is still a large room for further performance improvements. We hope our new benchmark and observations will facilitate future research.
\end{abstract}

%% file: sections/1_introduction.tex
\section{Introduction}
\label{sec:intro}

Vision-Language Models (VLMs)~\cite{radford2021learning,li2022blip,zheng2023marinegpt,li2023blip,zhu2023minigpt,stevens2024bioclip,yang2024biotrove,ziqiang2024marineinst} have achieved state-of-the-art results in a wide range of visual understanding tasks, including open-vocabulary object recognition~\cite{radford2021learning}, image captioning~\cite{li2022blip,li2023blip}, phrase grounding~\cite{peng2023kosmos,ren2024grounded} and interactive visual understanding~\cite{gpt4o2024}, because of their strong comprehension ability to align visual contents and natural-language description. The growing ability has motivated not only the general public but also domain research from a broad spectrum of scientific and industrial fields to adopt VLMs for domain applications, such as medical analysis~\cite{li2023llava}, mathematical computation~\cite{gao2023g}, and scientific research~\cite{lu2022learn}. 

In this work, we focus on the potential ability of powerful VLMs for marine understanding~\cite{kiefer20231st,ziqiang2024marineinst}, which is overlooked by existing research, but shares invaluable importance for protecting our ecosystem. The oceans, covering around 71\% of the area of our blue planet, play vital roles in different fields, making marine research non-negligible. Regarding the importance of marine research, they remain logistically difficult and expensive to observe. Though qualitative results of VLMs in general scenarios so far are encouraging~\cite{li2023seed, li2023seed2, li2024seed2plus}, quantitative evaluation is of great necessity to systematically evaluate and compare the abilities of various VLMs to conduct marine visual understanding. 

Directly applying the existing VLMs for detailed marine visual understanding is non-trivial, and there are still some significant challenges. First of all, the underwater conditions~\cite{tali,ziqiang2024marineinst} contain the non-trimmed background, lacking prior knowledge for obtaining a reliable and comprehensive marine understanding. Furthermore, strong performance on generic datasets does not guarantee decent accuracy in specialised settings, where data distribution shift, domain gaps, and the lack of domain-specific knowledge can severely degrade model reliability, leading to significant hallucination~\cite{zheng2024exploring}. We argue that the general-purpose evaluation dataset does not faithfully reveal the VLMs' capability in addressing domain-specific requirements (\emph{e.g.}, biologists favor the population/density estimation~\cite{ziqiang2024coralscop,ziqiang2025coralsrt,wong2025coralscop}, object counting~\cite{sun2023ioc}, and relationship summarization~\cite{han2025coralvqa}), as it rarely provides tailored tasks or authoritative ground truth for domain research. Consequently, existing research cannot effectively and reliably evaluate the performance of VLMs in handling marine understanding. 

To satisfy the need for marine evaluation, a representative and rigorous benchmark tailored to a domain application is indispensable for tracking methodological progress and selecting reliable models. Besides, evaluating the performance of VLMs in marine research will provide valuable insights into the flexibility of existing VLMS as a domain-specific AI assistant. However, there are a few attempts~\cite{palnitkar2023chatsim,zheng2023marinegpt} to comprehensively evaluate VLM for more advanced analysis, which requires domain-specific knowledge and expertise. The foregoing analysis implies that a domain-aware evaluation dataset should satisfy two criteria: 1) questions should demand specialised marine knowledge rather than common sense; 2) capability dimensions should be defined at a granularity that reflects specific domain requirements.

In this paper, we take all the above-discussed challenges into consideration, presenting the first large-scale marine VLM dataset and benchmark called \datasetname. Our \datasetname is a multi-task dataset (including diverse question/task formats) with 2,000 manually constructed high-quality image-based question-answering pairs from 7 task dimensions and 20 domain-specific capability dimensions as illustrated in Figure~\ref{fig:teaser}. To retain the quality of the constructed benchmark, we have formulated a rigorous pipeline shown in Figure~\ref{fig:workflow} to construct our dataset, which involves \textit{visual necessity testing} and \textit{domain expert verification}. Furthermore, to alleviate subjective grading and promote evaluation efficiency/reliability, we introduce detailed, scalable evaluation procedures to comprehensively assess VLM performance across multiple question formats. We have benchmarked \modelCounts existing SOTA VLMs on our \datasetname on the right of Figure~\ref{fig:teaser}, where the best model could only achieve 49.58\% accuracy. Substantial progress is still needed to enhance VLMs' performance on marine visual understanding. Our contribution can be summarized:
\begin{itemize}[leftmargin=*, noitemsep, topsep=0pt]
    \item We curate the first marine VLM benchmark with 2,000 high-quality image-based question-answering pairs, dedicated to marine analysis, enabling rigorous assessment of models on marine vision language tasks. 
    \item We have included the domain requirements/challenges into our benchmark construction, where 20 manually constructed capacity dimensions could comprehensively measure VLMs' ability for marine understanding.
    \item Our experimental results and observations reveal limitations of existing VLMs, demonstrating persistent challenges in spatial reasoning, precise localization, species identification, and ecological knowledge integration.
\end{itemize}

%% file: sections/2_literature_review.tex
\section{Related Work}
\label{sec:literature_review}

\noindent\textbf{VLMs}. The impressive performance of ChatGPT~\cite{chatgpt} and GPT-4~\cite{openai2023gpt4} has led to increasing attention to produce more powerful LLMs as an AI assistant. VLMs equip LLMs with the ability to receive visual content, and they have unveiled remarkable zero-shot image-text capabilities in a conversational format. Flamingo~\cite{alayrac2022flamingo} pioneered web-scale vision-language pretraining by bridging image and text models. BLIP~\cite{li2022blip,li2023blip} bootstraps vision-language pre-training from frozen pre-trained image encoders and frozen language decoders. Based on BLIP-2~\cite{li2023blip}, MiniGPT-4~\cite{zhu2023minigpt} proposed a projection layer to align pre-trained vision encoders to frozen LLMs (\emph{e.g.} Vicuna~\cite{chiang2023vicuna}), and exhibited respectable zero-shot image comprehension in dialogues. GPT-4V~\cite{gpt4o2024} showcased impressive general-purpose visual understanding and reasoning abilities. However, these VLMs may still make mistakes, especially for the domain-specific knowledge, since it is not specifically optimized on the reliable domain-specific corpus/knowledge. 

\noindent\textbf{Marine datasets}. Several datasets have recently been introduced for the marine domain. They provide insight into VLMs’ capacity for marine understanding.  MarineInst~\cite{ziqiang2024marineinst} emphasizes fine-grained perception by generating captions for individual object instances, while SeaFloorAI~\cite{NEURIPS2024_274de7d6} concentrates on geological and spatial knowledge through question answering. \datasetname aims to evaluate broader model capabilities across the marine domain.

 \noindent\textbf{Benchmarking VLMs}. Evaluating VLMs remains challenging due to the difficulty of assessing both their visual perception capabilities and their alignment with the inherently subjective and associative nature of human perception~\cite{fu2023mme,singh2023assessing}. To support systematic evaluation, multiple benchmarks have been introduced. For example, the SEED-Bench series~\cite{li2023seed, li2023seed2, li2024seed2plus} assesses VLMs through hierarchical tasks spanning a wide range of capabilities, while MMStar~\cite{chen2024we} and related work~\cite{goyal2017makingvvqamatter} expose risks of knowledge leakage, whereby models infer answers from contextual cues rather than visual input. Evaluation efforts have also been extended to specialized domains such as recommendation~\cite{zhou2023exploring}, medicine~\cite{royer2024multimedeval, li2023llava}, multilingual understanding~\cite{inoue2024heron}, and mathematics~\cite{zhang2024mathversedoesmultimodalllm, gao2023g}, highlighting their potential but also their limitations. Addressing the lack of resources for marine applications, we introduce \datasetname, the first dataset tailored to evaluate VLMs in marine-centric tasks, explicitly incorporating domain-specific requirements.
 

%% file: sections/3_methodology.tex
\section{\datasetname}
\label{sec:methodology}

\begin{figure*}
    \centering
    \includegraphics[width=\linewidth]{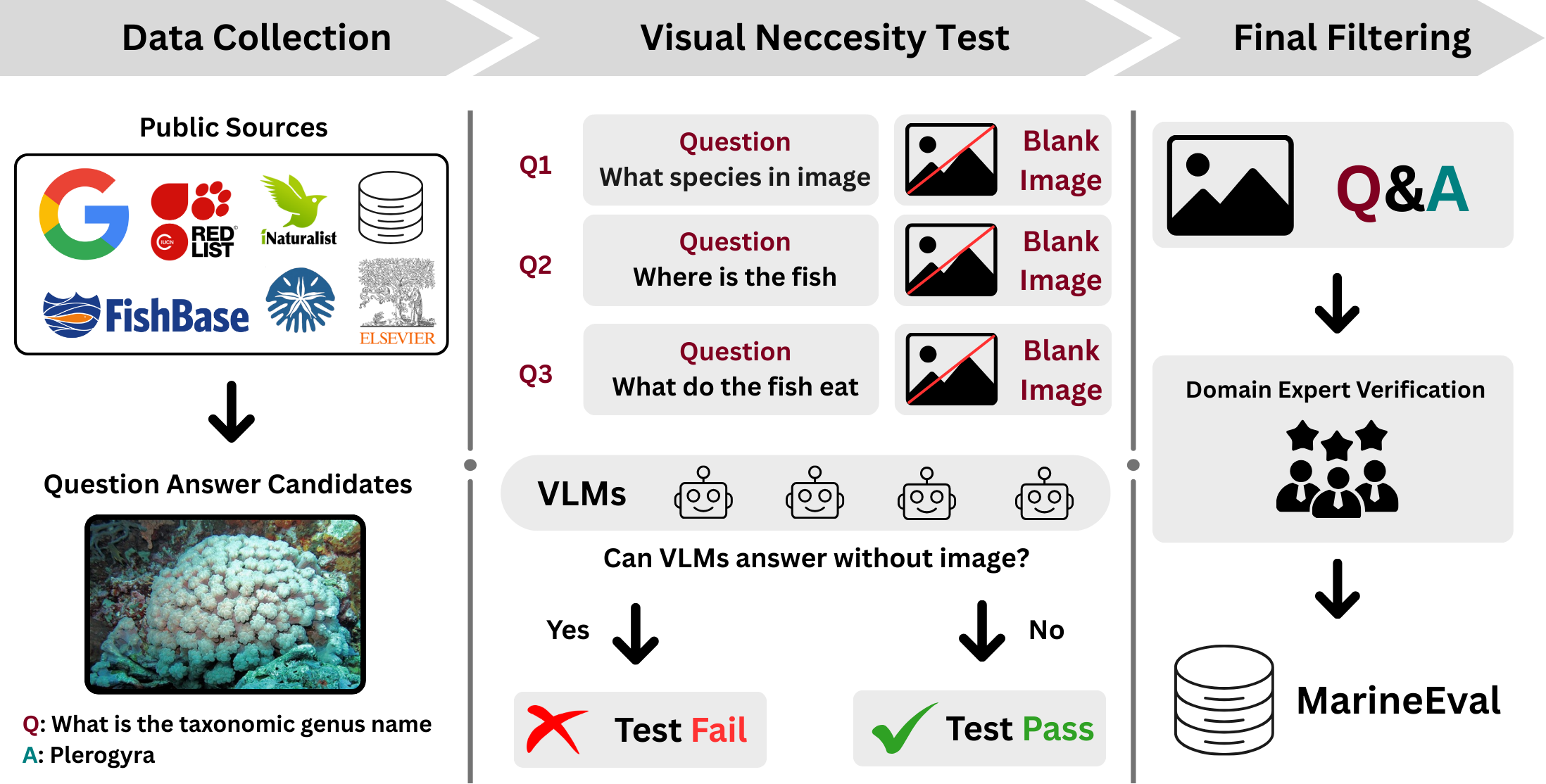}
    \caption{Workflow of \datasetname construction. 1) We first harvest diverse sources of candidate question-answer pairs, where the ground truth answer is post-processed by automatic programs or GPTs. 2) We adopt a visual necessity test to filter out pairs that are answerable without visual inputs. 3) Finally, domain experts construct and verify 2,000 high-quality pairs to constitute the final dataset.}
    \label{fig:workflow}
    \vspace{-0.2in}
\end{figure*}

In this section, we detail how we construct our \datasetname, which adheres rigorously to the evaluation criteria, while placing a strong emphasis on addressing specific marine challenges. It provides a tailored framework and well-defined capability dimensions to assess the effectiveness of VLMs in solving marine questions as illustrated in Figure~\ref{fig:workflow}. 

\subsection{General Criteria}
\label{sec:criteria}

\datasetname emphasizes 3 criteria for VLM evaluation. 

\noindent\textbf{Visual necessity}: VLMs should derive the answers based on visual content, rather than relying solely on the textual inputs. As highlighted in~\cite{chen2024rightwayevaluatinglarge, goyal2017makingvvqamatter}, there is a risk of knowledge leakage, where the question itself contains sufficient contextual cues for a well-trained LLM to get the correct answer. Such scenarios compromise the validity of the evaluation, as they fail to faithfully reflect the models' visual comprehension.

\noindent\textbf{Objectivity}: The evaluation rubrics should be clearly defined to avoid any subjective judgment. Existing works~\cite{inoue2024heron} adopt Likert-scale scoring, where the response is rated on a range (\emph{e.g.,} 1 to 5). Even though such scoring allows for more precise performance analysis, it may result in fluctuation across adjacent scores if the criteria are not clearly defined or the evaluators do not reach a common sense. This lack of clarity results in subjective and inconsistent evaluations, reducing the reliability of the results.

\noindent\textbf{Stability}: Evaluation results should demonstrate stability and consistency across repeated trials. However, existing approaches frequently depend on human annotators, whose judgments are inherently subjective and susceptible to cognitive bias. This reliance introduces variability in outcomes when different groups of evaluators conduct the same experiments, thereby undermining the reproducibility and comparability of the results.

With the consideration of the above criteria, we formulate our dataset construction (Section~\ref{sec:dataset_construction}) and evaluation process (Section~\ref{sec:settings}) accordingly.

\subsection{Dataset Construction}
\label{sec:dataset_construction}

MarineEval employs a systematic multi-step process for dataset construction, as shown in Figure~\ref{fig:workflow}:

\noindent\textbf{Data collection}: We first harvest a large range of diverse public datasets by aggregating and post-processing the corresponding visual annotations from these candidate sources, including public classification datasets~\cite{yang2024biotrove, vanhorn2018inaturalistspeciesclassificationdetection, oilSpillDataset, Ng_2022_CVPR, 9512396}, object detection datasets~\cite{10608133, Lamdouar20}, counting dataset~\cite{sun2023ioc}, marine-related books~\cite{Gosliner_Valdes_Behrens_2019, Humann_DeLoach_2010, Allen_2015}, scientific papers~\cite{MORROW2025104048,VENEGAS2025104053,FORGET2025104084,LIGNELL2025104049,scientific_paper1,scientific_paper2,scientific_paper3,scientific_paper4,scientific_paper6,scientific_paper7,scientific_paper8,scientific_paper9,scientific_paper10,scientific_paper11,scientific_paper12}, authoritative webpages~\cite{WoRMS20250718, fishbase, IUCN2022}, search engine, and private data. Detailed data source distribution is included in the Appendix.

\noindent\textbf{Visual necessity testing}: To eliminate questions that can be answered without visual content, a visual necessity test was conducted. Specifically, each question-answer pair in the candidate dataset was tested by removing the associated image and inputting it into five VLMs (Claude-3.7-Sonnet-Vision~\cite{anthropic2024claude3}, Gemini-2.0-Flash-Vision~\cite{gemini2024flash}, Grok-2-Vision~\cite{grok2024v2}, GPT-4o-Vision~\cite{gpt4o2024}, and Qwen-VL-Plus~\cite{Qwen-VL}). If any one of the models can infer the answer without relying on the visual content, the corresponding question will be deemed to exhibit data leakage and will be excluded from further consideration. We emphasize that such visual necessity testing could better fairly evaluate the ability of existing VLMs to truly understand the given visual contents. 

\noindent\textbf{Final filtering}. Besides the automatic and program-based construction, we also design a human-in-the-loop procedure to manually formulate the final dataset, where the ground truth answers are verified by domain experts.

\subsection{Evaluation Dimentions}

Our \datasetname could be systematically categorized into 7 overarching task dimensions and 20 capacity dimensions. The 7 tasks are summarized below, and some concrete examples are shown in Figure~\ref{fig:overview}. Details of 20 subfields are provided in the Appendix.
\begin{enumerate}
    \item \textbf{Species Comprehension} examines the capability of VLMs to identify and interpret species-level visual information, thereby contributing to biodiversity monitoring and ecological research.
    \item \textbf{Behavior \& Trait Extraction} focuses on the ability to derive meaningful insights into the behavior and physical traits of marine organisms, facilitating advancements in automated observational and documentary records.
    \item \textbf{Document Interpretation} evaluates the capacity of VLMs to analyze and derive insights from scientific literature and documentary sources. This functionality is especially critical for enhancing scientific understanding and generating insightful ecological reports.
    \item \textbf{Conservation \& Threat Analysis} emphasizes the ability of VLMs to accurately interpret domain-specific content, particularly in the context of endangered species and disaster classification.
    \item \textbf{Spatial Reasoning} measures spatial comprehension ability. While it is commonly evaluated in general scenarios, \datasetname specifically investigates whether VLMs sustain high performance in marine environments.
    \item \textbf{Marine Technology Understanding} evaluates understanding of marine technologies, which constitute a critical component of marine research.
    \item \textbf{Hallucination Resistance} tests the robustness of VLMs in avoiding erroneous or hallucinatory outputs. Specifically, it involves pairing generally true statements with images that depict corner cases or counterexamples to assess whether the VLM is susceptible to being misled by the accompanying statements.
\end{enumerate}

\subsection{Dataset Statistics and Specific Features}

\datasetname consists of 2,000 image-based question-answer pairs that span across 7 tasks and 20 capacity dimensions. To comprehensively evaluate the abilities of VLMs, we designed five distinct question formats: ``Yes-No questions'', ``multiple-choice questions'', ``localization questions'', ``closed-form questions'', and ``summarization questions'', as illustrated in Table~\ref{tab:question_format}. This diversity in question types enables \datasetname to assess a wide spectrum of capabilities for marine visual understanding, from basic factual judgement to complex reasoning and summarization tasks.

\begin{table}[t]
  \centering
  \small 
  \begin{tabular}{@{}p{2.5cm}p{5.3cm}@{}}
    \toprule
    \textbf{Question Format} & \textbf{Description} \\
    \midrule
    Yes-No Question & Models make binary classification to determine whether a statement is true or false. \\
    \midrule
    Multiple-Choice\newline Question & Models select one or more than one correct option from at least four choices. \\
    \midrule
    Localization\newline Question & Models are asked to provide bounding box of target objects in COCO format. \\
    \midrule
    Closed-Form\newline Question & Models respond in a restricted format (\emph{e.g.,} give a number or short phrases). \\
    \midrule
    Summarization\newline (Open-ended) & Models are asked to summarize the insight of the given image in free format. \\
    \bottomrule
  \end{tabular}
  \caption{Explanations of different question formats in \datasetname.}
  \label{tab:question_format}
\end{table}

Our dataset contains three key features compared with existing general-purpose benchmarks:

\noindent\textbf{1. Domain–specific marine knowledge requirements}. The majority of questions in \datasetname demand specialised expertise in marine science, such as \textit{taxonomic classification}, \textit{IUCN conservation status}, and \textit{biogeographic distribution} of specific organisms. This emphasis probes a knowledge space largely absent from mainstream training corpus, and thereby challenges the existing VLMs not only to retrieve and synthesize information but also to operate effectively within specialized knowledge domains. 

\noindent\textbf{2. Pronounced visual domain shift}. The collected images in \datasetname diverge markedly from the general-purpose dataset that focuses on common scenarios or human-centered events. Differently, \datasetname features a large proportion of underwater photographs, exhibiting low contrast, motion blur, colour attenuation, and a large range of perspectives. The images often capture complex habitats such as reef communities and pelagic schools, while some of the images are satellite imagery. These modalities create a substantial distribution shift and introduce visual complexities, thereby providing a robust test bed to stress the zero-shot visual generalisation of VLMs. 

\noindent\textbf{3. Practical evaluation setting with specific domain requirements}. \datasetname intentionally upheld both closed-form questions and open-ended questions to better represent real-world scenarios. While existing VLM datasets prioritize ease of evaluation by only providing ``Yes-No'' or ``multiple-choice questions'', \datasetname comprises 420 closed-form and open-ended questions (nearly one-fourth of the datasets) to measure models' ability to perform nuanced interpretation and free-form reasoning. Our design enables a more faithful measure of practical utility, where a fixed answer set is not available for a question.

\begin{figure*}
    \centering
    \includegraphics[width=0.9\linewidth]{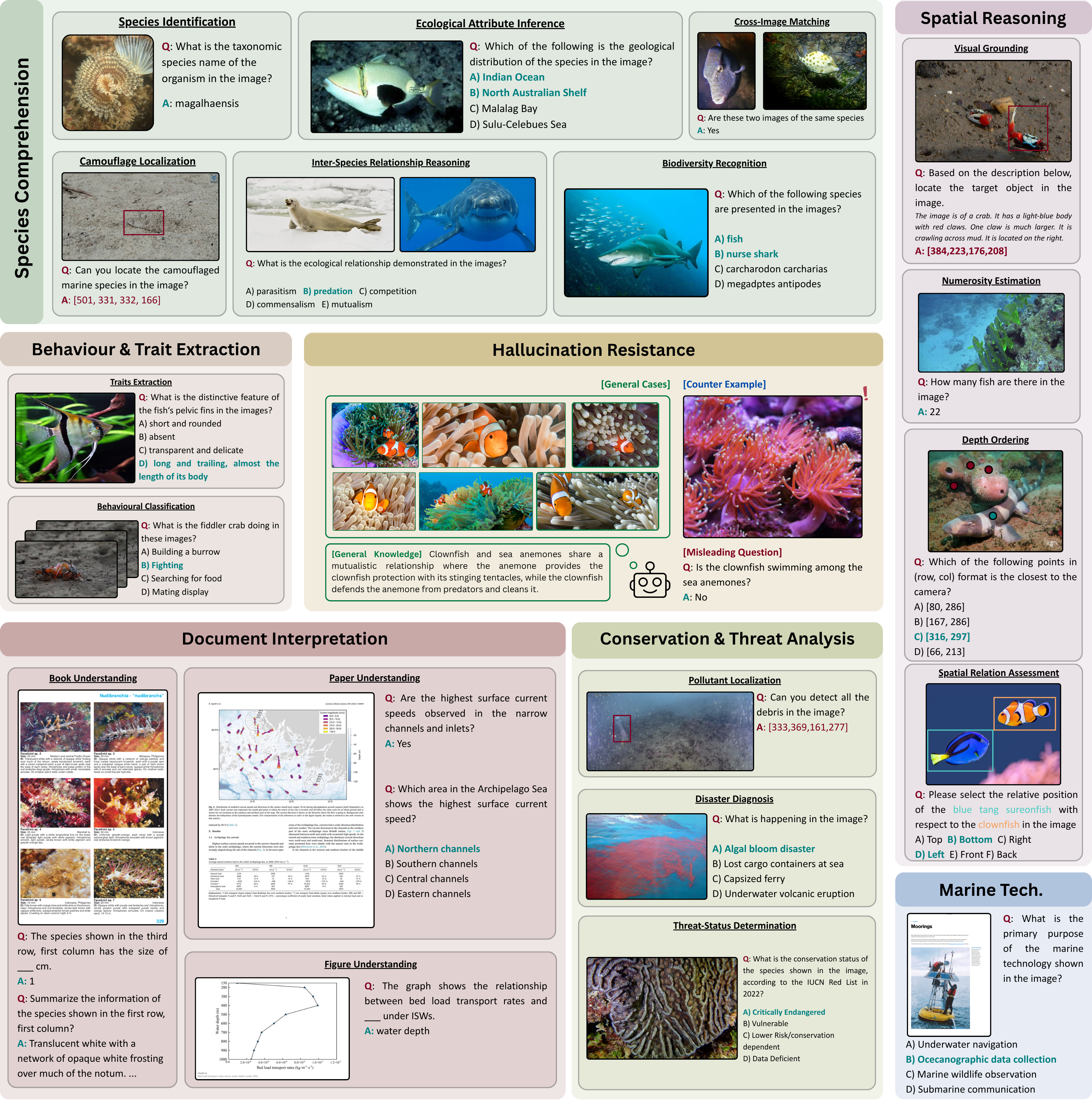}
    \caption{Overview of 7 task dimensions and 20 capacity dimensions of \datasetname. Best viewed in color.}
    \label{fig:overview}
    \vspace{-0.2in}
\end{figure*}

%% file: sections/4_experiments.tex
\section{Experiments}
\label{sec:experiments}
We first detail our experimental setting and then evaluate \modelCounts SOTA VLMs on \datasetname by conducting a quantitative analysis and summarizing key findings regarding limitations of VLMs in marine visual understanding.

\subsection{Experiment Settings}
\label{sec:settings}
We start by explaining how we evaluate the existing VLMs. To ensure \textit{objectivity}, \textit{stability}, and \textit{scalability}, as outlined in Section~\ref{sec:criteria}, we adopt a \textbf{binary judgement} evaluation strategy and report the final accuracy. To clearly verify the model responses, \datasetname classifies model outputs to either \textbf{correct} or \textbf{wrong}, regardless of their format or associated capability dimensions. Such an evaluation design introduces two benefits:

\begin{enumerate}
    \item \textbf{Clear marking rubrics}. Unlike Likert-scale scoring, where marking criteria can often be ambiguous, binary judgment lowers the evaluation difficulty by reducing the scoring task to a binary classification. It minimizes subjective interpretations and thus promotes greater objectivity and reproducibility in evaluation.
    \item \textbf{Easy comparison}. Binary judgement standardizes comparison by using \textit{accuracy} as a universal metric across different models and dimensions. By maintaining a consistent evaluation standard, comparisons across models and dimensions become more straightforward.
\end{enumerate}

We then provide the detailed evaluation metrics for computing the final accuracy regarding different question formats. For the ``Yes-No'' and ``Multiple-Choice'' questions, we first utilize the template matching to compute the accuracy of generated responses of various VLMs. During the evaluation procedure, we found some VLMs frequently violated the required response format, as these VLMs cannot strictly follow the user instructions to generate the required responses. To address this issue, we evaluate the models by appending each option to the question and computing the log-probability of the entire answer sequence. Instead of using raw logits, we compare the summed $\log p$ values over all tokens in the option. This avoids numerical issues and allows for fair comparison across options of varying lengths. The option with the highest total (or average) log-probability is selected as the final answer, and then we can eliminate the format issues for both ``Yes-No'' and ``Multiple-Choice'' questions. We adopt models' native decoding strategies for computing the probabilities.

\noindent\textbf{Localization Questions}. We evaluate the ability of VLMs to localize the specified object by the user instructions in the given image. We ask VLMs to follow the COCO format: \textit{$(x,y,w,h)$} to yield the bounding box prediction. Then we compute the Intersection-of-Union (IoU) between the prediction and ground truth. We regard the output with an IoU score over 0.3 as an accurate prediction. 

\begin{figure}
    \centering
    \includegraphics[width=\linewidth]{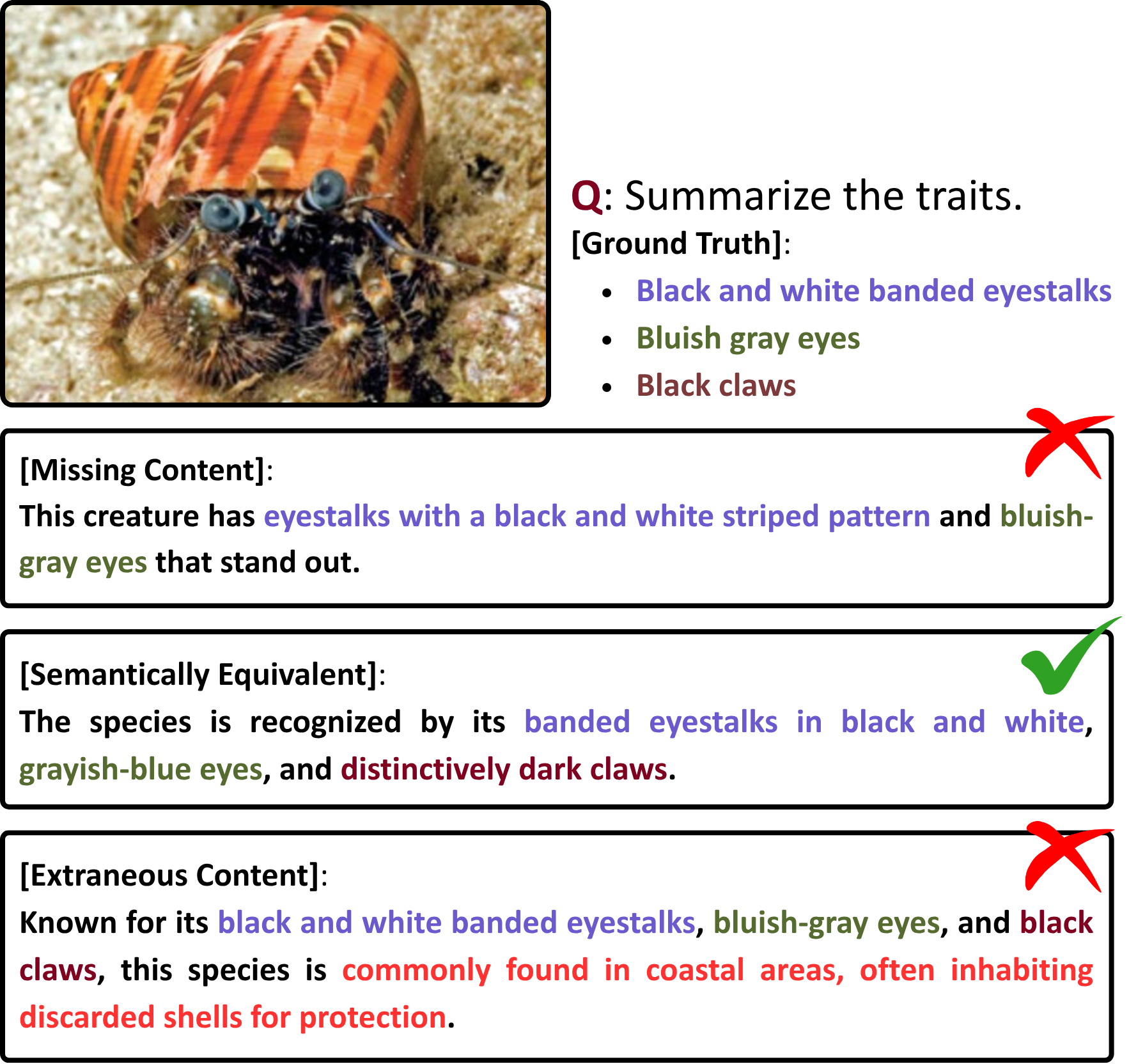}
    \caption{Open-ended responses frequently include omissions or irrelevant content, which hinders reliable evaluation. We employ the LLM for judgments to compare the ground truth answers and generated responses.}
    \label{fig:semantic_equavalent}
\end{figure}

\noindent\textbf{LLM for Judgement}. While existing datasets often use multiple-choice questions to reduce ambiguity, \datasetname includes complex closed-form and open-ended questions to better reflect real-world marine scenarios and enable more comprehensive VLM evaluation. Evaluating open-ended responses is challenging due to nuanced interpretation, as generated answers may differ syntactically yet match semantically with ground truth. We employ LLMs for automated accuracy assessment, enabling scalable semantic comparison between ground truth and candidate responses. In detail, for both closed-form and open-ended questions, we first construct ground truth answers by domain experts as illustrated in Figure~\ref{fig:semantic_equavalent}, where keypoints are summarized. Then we feed both the generated responses from the VLMs and the summarized keypoints by the humans to the LLMs to perform the matching from two aspects: whether there are 1) missing contents or 2) extraneous contents in the generated responses compared with the ground truth answers. In our experiments, considering there are some potential biases within the LLMs, we have chosen 3 powerful LLMs: GPT-4o-mini, Grok-3-mini, and DeepDeek-chat in our experiments.

\subsection{Baselines}
In our experiments, we have included 12 open-source VLMs: DeepSeek-VL-chat~\cite{lu2024deepseekvl}, OpenFlamingo~\cite{Alayrac2022FlamingoAV}, Mini-Monkey~\cite{huang2024mini}, InternVL-2.5~\cite{chen2024internvl}, LLaVA-1.6 Vicuna~\cite{liu2023llava}, InternLM-XComposer2.5~\cite{internlmxcomposer2_5}, LLaVA-Next~\cite{liu2024llavanext} and InternVL-2~\cite{chen2024internvl}; and 5 close-source VLMs: Claude-3.7-Sonnet-Vision~\cite{anthropic2024claude3}, Gemini-2.0-Flash-Vision~\cite{gemini2024flash}, Grok-2-Vision~\cite{grok2024v2}, GPT-4o-Vision~\cite{gpt4o2024}, and Qwen-VL-Plus~\cite{Qwen-VL} for evaluation. In detail, we adopt the official models released on Huggingface for the open-source model, where the model size ranges from 1.8B to 38B. All experiments for the open-source model are conducted using 6 NVIDIA GeForce RTX 4090 D. All the used prompts and hyperparameter configurations will also be released. For the closed-source models, we conduct evaluations by calling their official APIs, with a total of 10,000 inference calls (5 closed-source models for 2,000 questions) made to ensure robust and fair comparison.

\begin{table*}[]
\centering
\small
\scalebox{0.9}{
\begin{tabular}{l|c|ccccccccc}
\hline
\multicolumn{11}{c}{\textit{Open-source VLMs}} \\ \hline
\multicolumn{1}{l|}{Model} &
  \#. Params. &
  B\&TE &
  C\&TA &
  DI &
  HR &
  MTU &
  SR &
  SC &
  Avg. &
  Total\\
\hline
\multicolumn{1}{l|}{DeepSeek-VL-chat~\cite{lu2024deepseekvl}} &
  1.3B &
  27.86 &
  39.33 &
  11.00 &
  59.00 &
  34.31 &
  22.25 &
  18.33 &
  30.30 &
  24.96\\
\multicolumn{1}{l|}{OpenFlamingo~\cite{Alayrac2022FlamingoAV}} &
  2B &
  20.90 &
  40.33 &
  5.33 &
  60.00 &
  21.57 &
  8.25 &
  9.83 &
  23.74 &
  17.62\\
\multicolumn{1}{l|}{Mini-Monkey~\cite{huang2024mini}} &
  2B &
  44.28 &
  50.33 &
  33.00 &
  58.00 &
  74.51 &
  12.75 &
  27.67 &
  42.93 &
  34.45\\
\multicolumn{1}{l|}{InternVL-2.5~\cite{chen2024internvl}} &
  4B &
  65.17 &
  56.67 &
  54.00 &
  64.00 &
  \textbf{80.39} &
  16.75 &
  29.33 &
  52.33 &
  42.54\\
\multicolumn{1}{l|}{LLaVA-1.6 Vicuna~\cite{liu2023llava}} &
  7B &
  68.66 &
  52.00 &
  38.67 &
  53.00 &
  71.57 &
  \textbf{34.00} &
  37.33 &
  50.75 &
  44.73\\
\multicolumn{1}{l|}{InternLM-XComposer2.5~\cite{internlmxcomposer2_5}} &
  7B &
  64.18 &
  60.33 &
  49.33 &
  52.00 &
  75.49 &
  14.00 &
  30.17 &
  49.36 &
  41.14\\
\multicolumn{1}{l|}{LLaVA-Next~\cite{liu2024llavanext}} &
  8B &
  44.78 &
  \textbf{69.67} &
  25.67 &
  32.00 &
  54.90 &
  32.00 &
  26.67 &
  40.81 &
  37.54\\
\multicolumn{1}{l|}{InternVL-2~\cite{chen2024internvl}} &
  8B &
  55.22 &
  55.00 &
  46.00 &
  65.00 &
  78.43 &
  16.50 &
  34.17 &
  50.05 &
  41.44\\
\multicolumn{1}{l|}{InternVL-2.5 (26B)~\cite{chen2024internvl}} &
  26B &
  35.32 &
  41.67 &
  47.00 &
  66.00 &
  74.51 &
  25.00 &
  32.33 &
  45.98 &
  38.59\\
\multicolumn{1}{l|}{LLaVA-Next-Qwen~\cite{liu2024llavanext}} &
  32B &
  67.16 &
  60.00 &
  38.33 &
  65.00 &
  72.55 &
  16.50 &
  \textbf{43.67} &
  51.89 &
  44.78\\
\multicolumn{1}{l|}{LLaVA-1.6 Hermes-Yi~\cite{liu2023llava}} &
  34B &
  68.66 &
  52.00 &
  38.67 &
  53.00 &
  71.57 &
  \textbf{34.00} &
  37.33 &
  50.75 &
  44.73\\
\multicolumn{1}{l|}{InternVL-3~\cite{zhu2025internvl3}} &
  38B &
  \textbf{74.13} &
  48.33 &
  \textbf{60.33} &
  \textbf{68.00} &
  78.43 &
  22.50 &
  39.83 &
  \textbf{55.94} &
  \textbf{47.53}\\
\hline
\multicolumn{1}{l|}{Avg. across models} &
  $-$ &
  53.81 &
  52.77 &
  37.19 &
  59.42 &
  71.19 &
  21.23 &
  30.27 &
  46.14 &
  39.17\\
\hline
\multicolumn{11}{c}{\textit{Close-source VLMs}} \\ \hline
\multicolumn{1}{l|}{Model} &
  \#. Params. &
  B\&TE &
  C\&TA &
  DI &
  HR &
  MTU &
  SR &
  SC &
  Avg. &
  Total\\
\hline
\multicolumn{1}{l|}{Claude-3.7-Sonnet-Vision~\cite{anthropic2024claude3}} &
  $-$ &
  68.16 &
  53.67 &
  52.33 &
  71.00 &
  83.33 &
  24.50 &
  45.17 &
  56.88 &
  48.93\\
\multicolumn{1}{l|}{Gemini-2.0-Flash-Vision~\cite{gemini2024flash}} &
  $-$ &
  65.17 &
  \textbf{60.67} &
  \textbf{59.67} &
  \textbf{74.00} &
  \textbf{87.25} &
  29.00 &
  \textbf{55.33} &
  \textbf{61.59} &
  \textbf{55.07}\\
\multicolumn{1}{l|}{Grok-2-Vision~\cite{grok2024v2}} &
  $-$ &
  \textbf{77.61} &
  54.67 &
  27.33 &
  \textbf{74.00} &
  70.59 &
  \textbf{34.50} &
  54.00 &
  56.10 &
  50.42\\
\multicolumn{1}{l|}{GPT-4o-Vision~\cite{gpt4o2024}} &
  $-$ &
  69.15 &
  44.67 &
  51.67 &
  72.00 &
  62.75 &
  26.50 &
  40.50 &
  52.46 &
  45.58\\
\multicolumn{1}{l|}{Qwen-VL-Plus~\cite{Qwen-VL}} &
  $-$ &
  52.24 &
  41.00 &
  42.00 &
  71.00 &
  85.29 &
  25.00 &
  39.50 &
  50.86 &
  42.39\\
\hline
\multicolumn{1}{l|}{Avg. across models} &
  $-$ &
  66.07 &
  50.34 &
  46.20 &
  72.40 &
  77.64 &
  27.90 &
  46.10 &
  55.18 &
  48.08\\
\hline
\multicolumn{11}{c}{\textit{Human Performance}} \\ \hline
\multicolumn{1}{l|}{General Background} &
  $-$ &
  68.65 &
  54.33 &
  60.17 &
  82.00 &
  76.96 &
  51.50 &
  31.42 &
  60.72 &
  51.75\\
\multicolumn{1}{l|}{Marine Background} &
  $-$ &
  75.00 &
  70.33 &
  69.67 &
  83.00 &
  72.00 &
  64.00 &
  57.50 &
  70.31 &
  66.35\\ 
\hline
\end{tabular}
}
\caption{The average accuracy across 7 task dimensions. Abbreviation: Behavior \& Trait Extraction (B\&TE), Conservation \& Threat Analysis (C\&TA), Document Interpretation (DI), Hallucination Resistance (HR), Marine Technology Understanding (MTU), Spatial Reasoning (SR), Species Comprehension (SC). We calculate the average accuracy across 7 tasks and the total accuracy of 2,000 questions. $-$ indicates the number cannot be computed.}
\vspace{-0.2in}
\label{tab:overall}
\end{table*}

\subsection{Results and Observations}
\label{sec:results_and_observations}
The quantitative result comparisons between all the benchmarked VLMs are reported in Table~\ref{tab:overall}. We report the detailed accuracy of 7 task dimensions in our \datasetname, the average accuracy across these 7 tasks, and the total accuracy (the primary performance metric) on the total dataset. We also recruited participants from both general and marine backgrounds to answer the questions as a reference to upper-bound human performance. We have the following observations regarding the results:

\noindent\textbf{Inefficacious spatial and species understanding}. Spatial Reasoning (\SR) and Species Comprehension (\SC) remain among the most challenging capabilities for all evaluated models. Spatial reasoning tasks, such as image grounding and depth ordering, require fine-grained geometric representations that are insufficiently captured by current VLMs. Species comprehension, on the other hand, involves taxonomic identification and the inference of ecological attributes, which are beyond the scope of general-purpose VLMs. Further analysis on the impact of the domain gap (please refer to supplmentary) indicates that the limited performance in species comprehension is largely attributed to the models' lack of domain-specific knowledge. In contrast, the poor performance in spatial reasoning primarily stems from an inherent deficiency in spatial understanding in the general setting.
    
\noindent\textbf{Ecological insight scarcity}. The performance of Conservation \& Threat Analysis (\CTA) is still low for all models, \CTA questions involve disaster diagnostics and IUCN conservation status prediction, which represent corner cases and rare knowledge that is sparsely represented on the open web. Our results suggest that simply enlarging general-purpose corpora fails to cover long-tail ecological phenomena and the specialised reasoning they require.
    
\noindent\textbf{Model choice}. Model scale is not a reliable performance predictor of performing marine visual understanding. InternVL-2.5 (4B size) outperforms several larger models (even double-sized) and surpasses the closed-source models on multiple axes. This outcome underscores that architectural choices, vision encoders, and training strategy can outweigh parameter count. It also suggests diminishing returns for brute-force scaling when domain-specific supervision is scarce. 

\begin{table}[t]
  \centering
  \scalebox{0.65}{
    \begin{tabular}{@{}p{3.0cm}>{\centering\arraybackslash}p{3.0cm}>{\centering\arraybackslash}p{2.8cm}>{\centering\arraybackslash}p{2.8cm}@{}}
    \toprule
    Question Format & Acc. (w/ visuals) & Acc. (w/o visuals) & Random guessing\\
    \midrule
    Yes-No & 62.24 & 42.66 & 50.00 \\
    MCQ & 43.28 & 19.00 & 23.77 \\
    Localization & 01.27 & 00.00 & $-$ \\
    Closed-form & 20.85 & 04.34 & $-$ \\
    Summarization & 12.67 & 00.00 & $-$ \\
    Total & 35.84& 13.83& $-$ \\
    \bottomrule
  \end{tabular}
  }
  \caption{Effectiveness of visual inputs on \datasetname, where average accuracy of all 5 closed-source VLMs in each question format is reported. ``w/o visuals'' indicates the VLMs are not given with any visual input or given with a meaningless blank image. We also report the accuracy of random guessing if available.}
  \label{tab:visual_necessity_test}

\end{table}

\subsection{Further Analysis}
In this section, we provide more experimental analysis. 

\noindent\textbf{Visual Necessity}. We investigate the necessity of visual input in answering the questions, ensuring that the model cannot infer correct answers solely from the textual content. To this end, we conduct experiments under two settings: 1) \textit{with visuals}, where the actual visual input is provided, and 2) \textit{without visuals}, where a meaningless blank image is used as input. We evaluate all five closed-source VLMs and report their average accuracy across these settings. As shown in Table~\ref{tab:visual_necessity_test}, model performance declines substantially when visual information is removed. We also include the accuracy of random guessing as reference. Notably, a small subset of closed-form questions are answered correctly even without visual input, primarily in counting tasks where models’ random guesses coincide with the ground truth. Overall, these results confirm that the construction process of \datasetname does not unintentionally leak information to the VLMs, thereby ensuring the validity and integrity of the evaluation.

\noindent\textbf{LLM reliability}. We then investigate the reliability of using LLMs as judges, focusing on two key aspects: \textit{stability} and \textit{human alignment}. To assess stability, we repeat the evaluation procedure three times and report the mean and standard deviation in Table~\ref{tab:stability}. The experimental results show that incorporating LLMs does not introduce instability when evaluating the responses generated by VLMs under our experimental setup. For human alignment, we randomly sample 500 question–response pairs, which are independently evaluated by both human annotators and LLM judges. The results indicate that the final judgments produced by the LLMs achieve a \textit{95.40\% agreement rate} with human evaluators, demonstrating the reliability of using LLMs.

\noindent\textbf{Potential data contamination}. We acknowledge the potential data contamination issue in \datasetname, where some evaluation data might overlap with the training data of existing VLMs since this overlap could result in a biased or unfair comparison. Considering that all the benchmarked VLMs were optimized on diverse and extensive training corpora, it is inherently challenging to guarantee that all the testing data in \datasetname is entirely unseen by these models, as part of the data sources of \datasetname also come from publicly available datasets or public websites. Finally, it is important to emphasize that the primary objective of \datasetname is to explore the strengths and limitations of current VLMs in addressing marine challenges, not to fully address the data contamination issue. 

\begin{table}[t]
  \centering
  \scalebox{0.8}{
    \begin{tabular}{@{}p{3.0cm}>{\centering\arraybackslash}p{3.0cm}@{}}
    \toprule
    Question Format & Acc. ($\text{mean}_{\textcolor{blue}{\text{std}}}$) \\
    \midrule
    Closed-form & $20.83_{\textcolor{blue}{00.04}}$ \\
    Summarization & $12.67_{\textcolor{blue}{00.00}}$\\
    \bottomrule
    \end{tabular}
  }
  \caption{Mean and standard deviation of accuracy among 3 trials of using LLMs for measuring the generated responses from VLMs.}
  \label{tab:stability}
\end{table}

%% file: sections/5_conclusion.tex
\section{Conclusion and Discussion}
\label{sec:conclusion}
In this work, we investigated whether existing VLMs can serve as domain experts in marine understanding, a field demanding specialized knowledge and nuanced understanding. Through the construction of \datasetname, the first large-scale marine VLM benchmark encompassing 2,000 image-based QA pairs across 7 task dimensions and 20 capacities, we rigorously evaluated 12 open- and 5 closed-source VLMs. Our experiments revealed a critical gap: while general-purpose VLMs excel in broad tasks, they struggle with marine understanding and exhibit notable hallucinations when addressing spatial localization and species identification tasks.

%% file: sections/6_acknowledgement.tex
\section{Acknowledgement}

This project was partially supported by Bridging Horizons: An AI-Powered STEM Learning Initiative in Space and Marine Education under the EdUHK–HKUST Joint Centre for Artificial Intelligence, the HKUST Marine Robotics and Blue Economy Technology Grant, and the Marine Conservation Enhancement Fund (MCEF20107 and MCEF22112).

The authors would also like to express their sincere gratitude to the ``Sustainable Smart Campus as a Living Lab'' (SSC) program at HKUST for its vital support. The program and its dedicated staff not only contributed essential funding and coordination but also fostered the integration of sustainability into campus operations, providing a real-world demonstration of the principles that underpin this research.